\documentclass[letterpaper, 10 pt, conference]{ieeeconf}  

\IEEEoverridecommandlockouts                              
\usepackage{graphicx} 

\usepackage{amssymb}  
\usepackage{mathtools}          

\usepackage{amsmath}
\usepackage{array}
\usepackage{textcomp}

\title{\bf
Fully Convolutional Networks for Handwriting Recognition
}

\author{Felipe Petroski Such\textsuperscript{*}, Dheeraj Peri\textsuperscript{*}, Frank Brockler\textsuperscript{\textdagger}, Paul Hutkowski\textsuperscript{\textdagger}, Raymond Ptucha\textsuperscript{*} 
\\{\textsuperscript{*}Rochester Institute of Technology, USA}, {\textsuperscript{\textdagger}Kodak Alaris, USA}
\\{\{rwpeec, fps7806, dp1248\}@rit.edu}
}

\begin{document}

\maketitle
\thispagestyle{empty}
\pagestyle{empty}

\begin{abstract}

Handwritten text recognition is challenging because of the virtually infinite ways a human can write the same message. Our fully convolutional handwriting model takes in a handwriting sample of unknown length and outputs an arbitrary stream of symbols. Our dual stream architecture uses both local and global context and mitigates the need for heavy preprocessing steps such as symbol alignment correction as well as complex post processing steps such as connectionist temporal classification, dictionary matching or language models. Using over 100 unique symbols, our model is agnostic to Latin-based languages, and is shown to be quite competitive with state of the art dictionary based methods on the popular IAM and RIMES datasets. When a dictionary is known, we further allow a probabilistic character error rate to correct errant word blocks. Finally, we introduce an attention based mechanism which can automatically target variants of handwriting, such as slant, stroke width, or noise. 

\end{abstract}

\section{INTRODUCTION}

Convolution Neural Networks (CNN) have enabled many recent successes in pattern recognition systems. For example, deep CNNs have become ubiquitous in classification, segmentation and object detection. This success has also been demonstrated in Optical Character Recognition (OCR) systems, where CNNs can predict a sequence of characters from machine generated text with near-perfect accuracy.

Despite a shift towards digitized information exchange, there is still a need for handwritten inputs in many documents such as invoices, taxes, memos, and questionnaires.  Intelligent Character Recognition (ICR) is the task of deciphering digitized handwritten text. ICR is quite a bit more challenging than OCR because no two handwritten symbols are identical. ICR handwriting systems can be online or offline.  The former records stroke sequences say on a tablet, while the latter has no temporal information. This paper is concerned with offline ICR.  

Like OCR, ICR systems first extract lines of text and optionally can segment lines into word blocks separated by white space. ICR systems can feed these word blocks of symbols, often in context with surrounding lines of text, into lexicon-based classifiers.  Constraining the output to a lexicon of words can result in very accurate systems, but these same systems unfortunately cannot generalize to commonly occurring sequences such as addresses, phone numbers, and surnames.  Removal of dictionary lookup allows the recognition of unbounded dictionaries, but at the expense of decreased accuracy.


Jaderberg et al. \cite{jaderberg2014synthetic} proposed a framework which does word level recognition using CNNs on OCR tasks. Poznanski and Wolf \cite{poznanski2016cnn} used deep CNNs to extract n-grams which feed Canonical Correlation Analysis (CCA) for final word recognition from a fixed vocabulary.

A variant of Recurrent Neural Networks (RNNs), named Long Short Term Memory (LSTM) \cite{hochreiter1997long} has shown incredible progress in capturing long term dependencies and have been used in sequence prediction. Some works \cite{menasri2012a2ia,bluche2014comparison,pham2014dropout,doetsch2014fast} split an image into segments, then feed the sequence of segments into a RNN to predict a corresponding sequence of output characters. Connectionist Temporal Classification (CTC) \cite{graves2006connectionist} further eliminates the need for precise alignment. Xie et al. \cite{xie2016fully} used CNNs to feed a multi-layer LSTM network for handwritten Chinese character recognition. Similar techniques have also been used for text recognition in natural imagery \cite{shi2017end,shi2016robust}. 

A combination of convolutional networks and recurrent neural networks have been used by \cite{sun2016convolutional,voigtlaender2016handwriting} which use convolution layers for feature extraction and recurrent layers as sequence predictors. \cite{voigtlaender2016handwriting} performed ICR at the paragraph level to include language context.

Fully Convolutional neural Network (FCN) methods \cite{long2015fully,chen2016deeplab} take in arbitrary size images and output pixel level classification. Extracted portions of handwritten text have arbitrary length and can benefit from FCN methods. By using a CNN to estimate the number of symbols in a word block, word blocks can be resized to a canonical representation tuned to a FCN architecture. Knowing the average symbol width, a FCN model can perform accurate symbol prediction without CTC post processing. 

This paper proposes a method to obtain character based classification without relying on predefined dictionaries or contextual information. We believe our method is the first that can reliably predict both arbitrary symbols as well as words from a dictionary using a single architecture without any pre- or post-processing. The novel contributions of this paper are: 1) Introduction of a dual stream fully convolutional CNN architecture for accurate symbol prediction; 2) Usage of a probabilistic character error rate that calculates a word probability from a sequence of character probabilities; and 3) Creation of a difficult, but realistic block based dataset derived from the recently released NIST single character dataset \cite{nist19}.

\begin{figure*}
  \centering
 \centerline{\includegraphics[width=1\textwidth,]{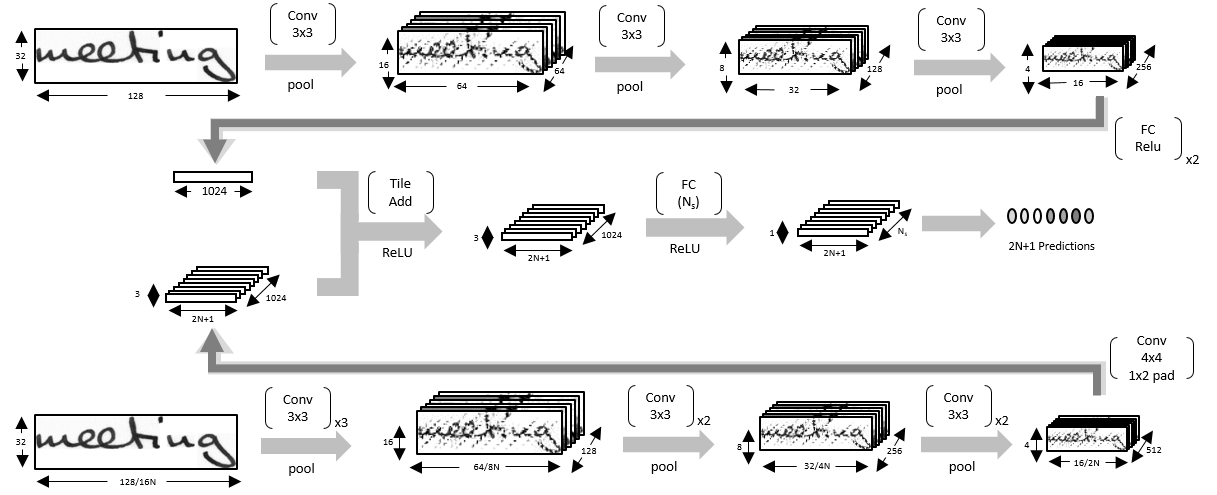}}
  \caption{The Symbol CNN is a fully convolutional model comprised of two networks. The bottom part of the network takes an input image of size $32 \times 128N$, where $N$ is the number of symbols in the word block. The top part of the network processes an image of size $32 \times 128$ and adds context during symbol prediction.}
\label{fig:icr_main}
\end{figure*} 
\section{Related Work}

Recent works in scene text recognition \cite{yadav2017deep,ahmed2017deep} use convolutional neural networks and pre-processing techniques like adjusting orientations to predict characters from a scene text. Batuhan et al. \cite{balci2017handwritten} proposed an end-to-end framework of recognizing characters from a document by using LSTMs for character segmentation and CNNs for character classification. 

One of the advantages of using CNNs is that inputs can be unprocessed data such as raw pixels of an image, rather than extracting specific features \cite{trier1996feature} or pen stroke grid values. 
Connectionist Temporal Classification (CTC) removes the need to forcefully align the input stream with character prediction locations \cite{graves2006connectionist}. One of the major advantages of the CTC algorithm is that you do not need properly segmented labeled data. Although the CTC algorithm takes care of the alignment of input with the output labels, it can be finicky to train and increases compute complexity during inference.

Huang and Srihari \cite{huang2008word} described an approach to separate a line of handwritten text to words. They proposed a gap metrics based approach to perform word segmentation task. 
Rather than segmenting words, Gader et al. \cite{gader1997handwritten} proposed character segmentation utilizing information as you move from background pixels to foreground pixels in horizontal and vertical directions of the character image. 

Doetsch et al. \cite{doetsch2014fast} proposed hybrid RNN-HMM for English offline handwriting recognition. 
They introduced a new variant of a LSTM memory cell by using a scalar multiple for every gate in each layer of the RNN. The scaling technique for LSTM gates reduced Character Error Rate (CER) by 0.3\%. Bluche et al. [1] compared CNN and traditional feature extraction techniques along with HMM for transcription.
Pham et al. \cite{pham2014dropout} proposed Multidimensional RNN using dropout to improve offline handwriting recognition accuracy by 3\%.
 
Dewan and Srinivasa \cite{dewan2012system} and Xie et al. \cite{xie2016fully} used CNNs for offline character recognition of Telugu and Chinese characters respectively. \cite{dewan2012system} used auto encoders, where the model was trained in a greedy layer wise fashion to learn weights in an unsupervised fashion, then fine-tuned with supervised data. 

\section{Methods}
An important part of character recognition is Region of Interest (ROI) extraction. Areas of text are extracted from documents using regional based classifiers such as R-CNNs \cite{ren2015faster} or pixel based segmentation \cite{chen2016deeplab}. These regions are split into lines of text using modified XY tree \cite{cesarini1999structured}. 
Each line of text is then split into word blocks, where a word block is a string of symbols separated by white space. The string of symbols could be a word, phone number, surname, acronym, etc. Word block boundaries can again be determined by modified XY tree.  
Using punctuation detectors \cite{silla2004analysis}, it is possible to detect and store a string of word blocks to form sentences, and possible to detect and store a string of sentences to form paragraphs. The FCNs in this research take extracted word blocks as input. Preprocessing such as contrast normalization and deslanting have shown to be effective for handwriting recognition \cite{kozielski2013improvements} \cite{voigtlaender2016handwriting}, but are not used in this research.

We propose a three staged approach in this paper for the recognition of handwritten characters. In the first stage, we train a CNN to quickly predict the word label for common words such as ``the'', ``her'', ``this'', etc. This model is a typical image classification model where the ground truth classes correspond to discrete words from the training set. We call this a Vocabulary CNN, as it can only predict the common lexicon of words from the training set. The architecture of the Vocabulary CNN is C(64,3,3)-C(64,3,3)-C(64,3,3)-P(2)-C(128,3,3)-C(128,3,3)-C(256,3,3)-P(2)-C(256,3,3)-C(512,3,3)-C(512,3,3)-P(2)-C(256,4,16)-FC(V)-SoftMax where C(D,H,W) stands for convolution with the dimensions of the filter as $H{\times}W$ and the depth $D$. Each convolutional layer is followed by a batch norm and ReLU layer. P(2) represents a 2$\times$2 pooling layer with stride 2. The last layer FC(V) is a fully connected layer with V being the lexicon size. We consider a prediction to be valid if the confidence level is more than 0.7, else we process the block through the next two stages.

In the second stage, we train a CNN to predict the number of symbols in a word block. We refer to this CNN as the Length CNN. This model has a similar architecture to the Vocabulary CNN with additional max-out layers in between convolutional layers. The ground truth classes correspond to the length of the word blocks in the training set. The predicted number of symbols from the Length CNN is used to resize the word block to a canonical representation of $32\times128N$, where $N$ is the number of symbols. This resized word block is then passed into the third stage which is a fully convolutional variant of CNN.

In the third stage, a dual stream FCN predicts an arbitrary length sequence of characters from a variable length word block. FCN's are a natural choice for processing variable length word blocks as regular CNNs require fixed length inputs. We call this third stage the Symbol CNN, as it can recognize words at a symbol level. Figure \ref{fig:icr_main} pictorially shows the dual stream architecture used. The top stream takes in an image of size $32 \times 128$ which is passed through a series of convolutional layers and a fully connected layer. The fully connected output (fc) size of the top stream is 1024. These features represent the global information in the word block. The second stream consists of only convolutional and pooling layers. The fully connected layer is replaced by a fully convolutional layer which makes $2N+1$ predictions for a word block of length $N$. The outputs from both the streams are added and passed through a fully connected layer of size $N_{s}$ where $N_{s}$ represents the lexicon of symbols. The final output size is $(2N+1) \times N_{s}$. We found that this dual stream model is able to encode rich information about the global and local context of the symbols in the word block.

\begin{figure}
  \centering
 \centerline{\includegraphics[width=0.45\textwidth,]{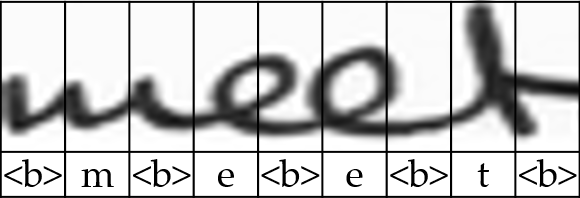}}
  \caption{During training, blank space characters(\textless b \textgreater) are added in between ground truth symbols to improve symbol alignment.}
\label{fig:gt_align}
\end{figure}

$2N+1$ predictions were chosen such that each of the $N$ symbols in a word block are straddled on both sides by a small gap which we refer to as a blank space character. In particular we align the ground truth symbols with blank spaces as shown in Figure \ref{fig:gt_align}. In our experiments, we found this to be intuitive and empirically performs well in both dictionary constrained and non-dictionary use cases. There are a total of 123 ground truth symbols which include alphabets (English, French, Spanish, German), numbers, and special characters such as \$, \&, period etc. 

By not constraining the Symbol CNN output, our model is able to predict any random sequence of characters such as phone numbers, social security numbers, email ID etc. As such, the Symbol CNN is particularly useful to digitize tax, insurance, and medical documents.  The dual stream and full convolutional nature of the model enables it to be quite effective at symbol recognition with or without a word lexicon. 

\section{CER and Vocabulary Matching}
	\begin{figure}
  \centering
 \centerline{\includegraphics[width=0.5\textwidth,]{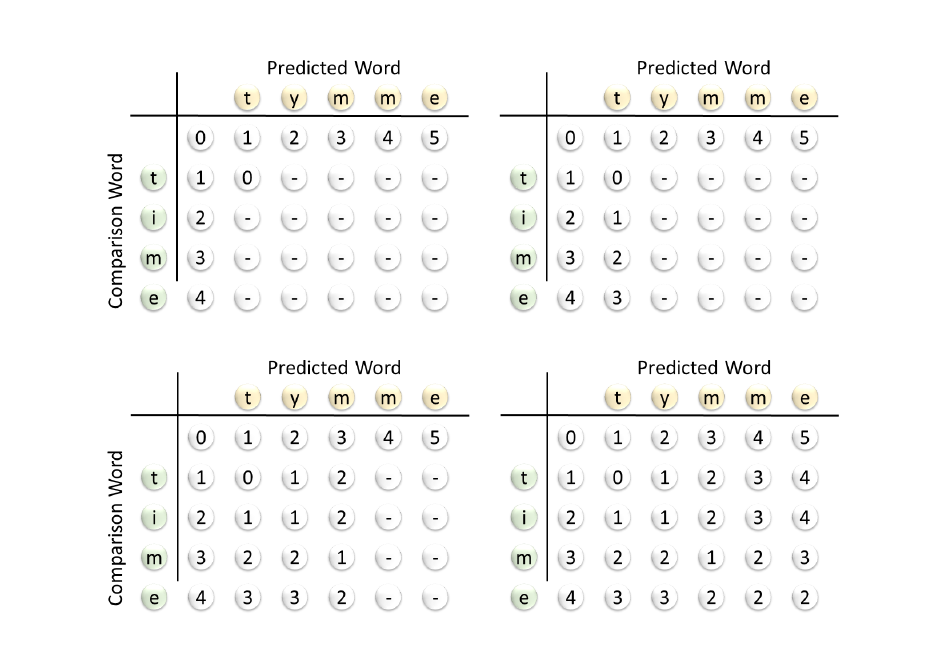}}
  \caption{Calculating CER error between words \textit{tymme} and \textit{time} using dynamic programming. From left to right: after one step, after finishing ``t'', after finishing first ``m'', and final
CER of ’2’.}
\label{fig:dynamic}
\end{figure}   
We report our results using Character Error Rate (CER): 
\begin{equation}
    CER = R + D + C
\end{equation}

where $R$ is number of characters replaced, $D$ is the number of characters deleted and $C$ is the number of correct characters. This can be efficiently computed using dynamic programming and we show an example of the computation in Figure \ref{fig:dynamic}. Equation \ref{cer_dynamic} describes the CER computation where $C_{0,0} = 0$ and CER = $C_{l,h}$ where $l$ is the length of the label and $h$ is the length of the prediction. $p_{i}$ is the $i^{th}$ character of the prediction and $L_{i}$ is the $j^{th}$ character of the label.
\begin{equation}
    C_{i,j} = \textit{min($C_{i-1,j}$+\textnormal{1}, $C_{i, j-1}$+\textnormal{1}, Diag)}
\label{cer_dynamic}
\end{equation}
\hspace{2mm}   where: 
\[Diag =  \begin{cases} 
       C_{i-1, j-1}, & if \hspace{1mm} p_{i} = L_{j} \\
       C_{i-1, j-1} + 1, & otherwise \\
   \end{cases}
\]

To improve performance in applications that have a known-limited vocabulary, we applied a CER-based vocabulary matching system using dynamic programming as shown in (\ref{vocab_match}), 

\begin{equation}
    W(p) = \underset{L{\in}V}{\textrm{arg min }}\textit{CER(p,L)}
\label{vocab_match}
\end{equation}

where $V$ is the vocabulary set and ${W(p)}$ is the word prediction based on the sequence prediction $p$. The sequence prediction $p$ is the set of symbols predicted by FCN.

The above method improved CER, but discards
most of the information computed with the neural
network. An improvement to the above vocabulary matching system, which we refer to as probabilistic CER, uses character probabilities instead of the discrete top character prediction. Equation\ref{prob_cer} describes probabilistic CER method.
      
\begin{align}
	\label{prob_cer}
    C_{i,j} = \textit{min}(C_{i-1,j}+1 - P(p_{i} = L_{j}),\nonumber\\
   C_{i,j-1} + 1- P(p_{i} = blank), \\
    C_{i-1,j-1} + 1- P(p_{i} = L_{j}),\nonumber)
\end{align}

Since we now have a method to compute word probabilities from sequence probabilities, we can combine predictions into the same system. We use the frequency of occurrence of a given word $C(L)$ to further improve the vocabulary matching using (\ref{freq}).
       
\begin{equation}
    W(p) = \underset{L{\in}V}{\textrm{arg min }}\textit{CER(p, L)} + \frac{1}{1+C(L)}
\label{freq}
\end{equation}
        
\section{Attention Modeling}
When known variation of handwriting is both common and easily computer generated, data augmentation can be used to generate a family of resized word blocks to predetermined conditions.  For example, given an input handwritten word, deslanting algorithms can be used to create a family of inputs, each at a predetermined slant. A similar family of inputs can be created by varying stroke width, noise variation, and paper/background.  Given a family of resized word blocks, a vector of attention weights associated with the family of resized word blocks is learned.  Using the generated vector attention weight, a single resized word block is formed as a linear combination of resized word blocks, and passed through the CNN in normal fashion. 

\section{Training}
We used Caffe \cite{jia2014caffe} to perform all experiments. Custom layers were implemented in python to handle ground truth and the blankspace symbol alignment. We trained jointly on IAM \cite{marti2002iam}, RIMES \cite{augustin2006rimes} and NIST \cite{nist19} datasets. All experiments used batch size of 64 (24 IAM, 24 RIMES and 16 NIST samples) and ran for 50k iterations. We used a learning rate of 0.001 to train the network with 0.9 momentum. The learning rate was decreased to 0.0001 after 40k iterations. We also used L2-regularization with $\lambda$ = 0.0025. 

\section{Results}
Results are demonstrated on the IAM, RIMES, and NIST offline handwritten datasets. The IAM \cite{marti2002iam} dataset contains 115,320 English words, mostly cursive, by 500 authors. This dataset includes training, validation, and test splits, where an author contributing to a training set, cannot occur in the validation or test split. The RIMES \cite{augustin2006rimes} dataset contains 60,000 French words, by over 1000 authors. There are several versions of the RIMES dataset, where each newer release is a super-set of prior releases. We utilize the popular ICDAR2011 release.
        
The NIST Handprinted Forms and Characters Database, Special Database 19 \cite{nist19}, contains NIST’'s entire corpus of training materials for handprinted document and character recognition. Each author filled out one or more pages of the NIST Form-based Handprint Recognition System. It publishes handprinted sample forms from 810,000 character images, by 3,600 authors.
        
\subsection{IAM Results}
We first test our system on the IAM English handwritten dataset. We fine-tuned our model on IAM and it achieves CER of 4.43\%. Table \ref{IAM_res} highlights the results of our model with dictionary correction which is quite competitive to the current leaders of this dataset. Table \ref{iam_sym} shows examples of predictions obtained on the IAM dataset using only the FCN model without dictionary correction.
  
\begin{table}[!ht]
\centering
\caption{Comparison of results on IAM dataset to previous methods.}
\label{IAM_res}
\begin{tabular}{|c|c|c|}
\hline
Model & WER & CER\\
\hline
Dreuw et al. \cite{dreuw2011hierarchical} & 18.8 & 10.1\\
\hline
Boquera et al. \cite{espana2011improving}& 15.5 & 6.90\\
\hline
Kozielski et al. \cite{kozielski2013improvements}& 13.30 & 5.10\\
\hline
Bluche et al. \cite{bluche2014comparison}& 11.90 & 4.90\\
\hline
Doetsch et al. \cite{doetsch2014fast}& 12.20 & 4.70\\
\hline
Our work & 8.71 & 4.43\\
\hline
Voigtlaender et al. \cite{voigtlaender2016handwriting}& 9.3 & 3.5\\
\hline
Poznanski and Wolf \cite{poznanski2016cnn}& \textbf{6.45} & \textbf{3.44}\\
\hline
\end{tabular}
\end{table}
       
Drewu et al. \cite{dreuw2011hierarchical} showed that competitive results can be obtained by hybrid approaches of MLP and Gaussian HMMs. Kozielski et al. \cite{kozielski2013improvements} used a novel HMM based system. Drewu et al. \cite{dreuw2011hierarchical} and Boquera et al. \cite{espana2011improving} use a hybrid neural network and Hidden Markov Model HMM approach. Bluche et al. \cite{bluche2014comparison} used Gaussian HMMs to initialize neural networks and showed that both deep CNNs and RNNs could produce state of the art results. Doetsch et al. \cite{doetsch2014fast} uses a custom LSTM topology along with CTC alignment. \cite{doetsch2014fast,bluche2014comparison} used all words in a sentence and paragraph respectively to provide word context. Poznanski and Wolf \cite{poznanski2016cnn} used deep CNNs to extract n-gram attributes which feed CCA word recognition. \cite{kozielski2013improvements,poznanski2016cnn,doetsch2014fast,espana2011improving} use deslanting, training augmentation, and an ensemble of test samples.

Our work uses a first Vocabulary CNN of 1100 common words. The Symbol CNN uses 123 symbols, and we use probabilistic CER correction. Aside from the probabilistic CER correction, no CTC alignment or CCA post correction was applied. Although our competitive results are not ranked the best, our processing path can work at both the symbol 
and lexicon level, and we include substantially more symbols than prior methods (e.g. \cite{poznanski2016cnn} can only recognize upper and lower case Latin alphabet).

\begin{table}[!ht]
  \centering
  \caption{Symbol sequence prediction on the IAM dataset. The third example has a questionable ground truth and a prediction that could be considered valid out of context.}
  \label{iam_sym}
  \begin{tabular}{ | c | c | c | }
  \hline
  Input & Label & Prediction \\ \hline
  	\begin{minipage}{0.1\textwidth}
      \includegraphics[width=\linewidth, height=6mm]{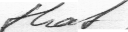}
    \end{minipage}
    & that & that
  \\ \hline
  \begin{minipage}{0.1\textwidth}
      \includegraphics[width=\linewidth, height=6mm]{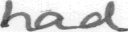}
    \end{minipage} & had & had
  \\ \hline
    \begin{minipage}{0.1\textwidth}
      \includegraphics[width=\linewidth, height=6mm]{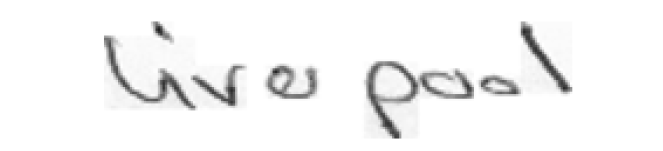}
    \end{minipage} & Liverpool & livepool
  \\ \hline
  \begin{minipage}{0.1\textwidth}
      \includegraphics[width=\linewidth, height=6mm]{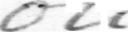}
    \end{minipage} & on & oui
  \\ \hline
  \begin{minipage}{0.1\textwidth}
      \includegraphics[width=\linewidth, height=6mm]{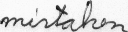}
    \end{minipage} & mistaken & mistahon
  \\ \hline
  \begin{minipage}{0.1\textwidth}
      \includegraphics[width=\linewidth, height=5mm]{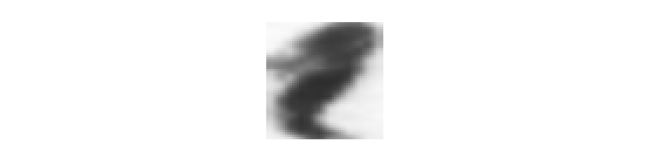}
    \end{minipage} & ' & ,
  \\ \hline
  \begin{minipage}{0.1\textwidth}
      \includegraphics[width=\linewidth, height=6mm]{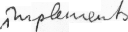}
    \end{minipage} & implements & implement
  \\ \hline
  \begin{minipage}{0.1\textwidth}
      \includegraphics[width=\linewidth, height=6mm]{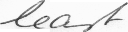}
    \end{minipage} & least & least
  \\ \hline
  \begin{minipage}{0.1\textwidth}
      \includegraphics[width=\linewidth, height=6mm]{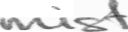}
    \end{minipage} & mist & mist
  \\ \hline
  \begin{minipage}{0.1\textwidth}
      \includegraphics[width=\linewidth, height=6mm]{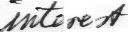}
    \end{minipage} & interest & interest
  \\ \hline
\end{tabular}
\end{table}

\subsection{RIMES Results}
We use a vocabulary CNN of 800 common words and fine-tuned our model on RIMES dataset. Table \ref{RIMES_res} shows our model obtained a CER of 2.22\% using an ensemble of three models, each with dictionary correction. Table \ref{rimes_sym} shows examples of predictions obtained from one of the models without dictionary correction. 
In general, errors can be attributed to character ambiguity, segmentation artifacts (sample ``effet'' contains a comma even though it isn\textquotesingle t part of the label).
For most of the examples in Table \ref{rimes_sym}, our  model made perfect predictions.

\begin{table}[!ht]
\centering
\caption{Comparison of results on RIMES dataset to previous methods.}
\label{RIMES_res}
\begin{tabular}{|c|c|c|}
\hline
Model & WER & CER\\
\hline
Kozielski et al. \cite{kozielski2013improvements}& 13.70 & 4.60\\
\hline
Doetsch et al. \cite{doetsch2014fast}& 12.90 & 4.30\\
\hline
Bluche et al. \cite{bluche2014comparison}& 11.80 & 3.70\\
\hline
Our work & 5.68 & 2.22\\
\hline
Poznanski and Wolf \cite{poznanski2016cnn}& \textbf{3.90} & \textbf{1.90}\\
\hline
\end{tabular}
\end{table}

\begin{table}[!ht]
  \centering
  \caption{Symbol sequence prediction on the RIMES dataset. }
  \label{rimes_sym}
  \begin{tabular}{ | c | c | c | }
  \hline
  Input & Label & Prediction \\ \hline
  	\begin{minipage}{0.1\textwidth}
      \includegraphics[width=\linewidth, height=5mm]{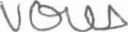}
    \end{minipage}
    & vous & vous
  \\ \hline
  \begin{minipage}{0.1\textwidth}
      \includegraphics[width=\linewidth, height=4mm]{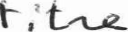}
    \end{minipage} & titre & titre
  \\ \hline
    \begin{minipage}{0.1\textwidth}
      \includegraphics[width=\linewidth, height=5mm]{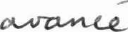}
    \end{minipage} & avanc\'e & avance
  \\ \hline
  \begin{minipage}{0.1\textwidth}
      \includegraphics[width=\linewidth, height=5mm]{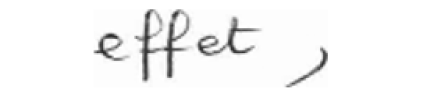}
    \end{minipage} & effet , & effett
  \\ \hline
  \begin{minipage}{0.1\textwidth}
      \includegraphics[width=\linewidth, height=5mm]{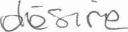}
    \end{minipage} & d\'esire & di\'esiire
  \\ \hline
  \begin{minipage}{0.1\textwidth}
      \includegraphics[width=\linewidth, height=5mm]{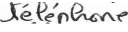}
    \end{minipage} & t\'el\'ephone & t\'el\'enhone
  \\ \hline
  \begin{minipage}{0.1\textwidth}
      \includegraphics[width=\linewidth, height=5mm]{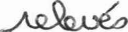}
    \end{minipage} & relev\'es & relves
  \\ \hline
  \begin{minipage}{0.1\textwidth}
      \includegraphics[width=\linewidth, height=5mm]{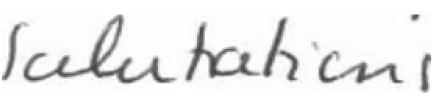}
    \end{minipage} & salutations & salutations
  \\ \hline
  \begin{minipage}{0.1\textwidth}
      \includegraphics[width=\linewidth, height=5mm]{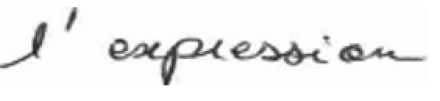}
    \end{minipage} & l'expression & l'expression
  \\ \hline
  \begin{minipage}{0.1\textwidth}
      \includegraphics[width=\linewidth, height=5mm]{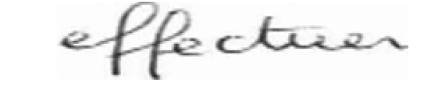}
    \end{minipage} & effectuer & effectuer
  \\ \hline
\end{tabular}
\end{table}

\begin{table}[!ht]
  \centering
  \caption{Symbol sequence prediction on NIST generated samples.}
  \label{nist_sym}
  \begin{tabular}{ | c | c | c | }
  \hline
  Input & Label & Prediction \\ \hline
  	\begin{minipage}{0.1\textwidth}
      \includegraphics[width=\linewidth, height=5mm]{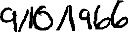}
    \end{minipage}
    & 9/10/1966 & 9/10/1966
  \\ \hline
  \begin{minipage}{0.1\textwidth}
      \includegraphics[width=\linewidth, height=5mm]{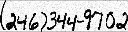}
    \end{minipage} & (246)344-9702 & (246)344-9702
  \\ \hline
    \begin{minipage}{0.1\textwidth}
      \includegraphics[width=\linewidth, height=5mm]{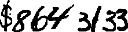}
    \end{minipage} & \$8643133 & \$8643133
  \\ \hline
  \begin{minipage}{0.1\textwidth}
      \includegraphics[width=\linewidth, height=5mm]{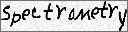}
    \end{minipage} & Spectrometry & Spectrometry
  \\ \hline
  \begin{minipage}{0.1\textwidth}
      \includegraphics[width=\linewidth, height=5mm]{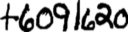}
    \end{minipage} & +6091620 & +6091620
  \\ \hline
  \begin{minipage}{0.1\textwidth}
      \includegraphics[width=\linewidth, height=5mm]{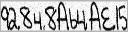}
    \end{minipage} &  92.84.8A.b4.AE.15 & 92.84.8A.b4AE.15
  \\ \hline
\end{tabular}
\end{table}

\subsection{NIST}

While there are several class specific handwritten datasets, both at the character and word level, there is no large handwritten dataset that concentrates on word blocks of arbitrary symbols. To test the performance of our model on generic word blocks made of arbitrary symbols, we created a new symbol recognition dataset by stochastically combining the NIST individual character images into realistic word blocks. Images of hand printed text are simulated by extracting character images from a randomly selected writer in the NIST dataset and concatenating them into word blocks of random dictionary words, random strings of alphanumeric characters, or random strings of numeric characters. In addition, the NIST dataset has been supplemented with handwritten punctuation, mathematical symbols and other common special characters such as the dollar sign, email character and the ampersand to facilitate in generating word block images of common form-field inputs. 
      
The images are further augmented by adding random amounts of displacement, stretching and rotation to each symbol to simulate the natural variability in writer's penmanship. A random amount of skew is then applied to each concatenated image to vary the slant of the word block. Finally, random amounts of noise and blur are added to simulate the effects of image digitization. 
Table \ref{nist_sym} shows randomly generated samples from NIST characters and corresponding predictions. 
Our model achieves 92.4\% accuracy on a subset of 12,000 word blocks (consisting English, French and special characters) generated from NIST. 

Some of the characters in handwritten word blocks are too hard to decipher and require contextual information for human or machine interpretation. In order to provide more contextual information and recognize characters which are ambiguous due to incomplete pen strokes and inconsistencies in handwriting, we utilized both the contextual path introduced at the top of Figure \ref{fig:icr_main} as well as an attention mechanism. After preprocessing an input word block with a series of known variations such as slant and noise, the attention mechanism formulates a single input as a linear combination of these variations. Although the attention mechanism is intuitive, empirical results have yet to show any significant improvement and were not used in the results in Tables \ref{IAM_res} or \ref{RIMES_res}.

\section{Conclusion}

We introduce a novel offline handwriting recognition algorithm using a fully convolutional network. Unlike lexicon constrained methods, our method can recognize common words as well as infinite symbol blocks such as surnames, phone numbers and acronyms. Our dual stream architecture along with blank space symbol alignment eliminates the need of complex character alignment methods such as CTC in recurrent based methods. Our fully convolutional model enables processing of arbitrary length inputs and utilizes a large symbol set.  Despite not using a word lexicon and handling of a large symbol set, our method performs on par with current state of the art methods on English based IAM, French based RIMES, and NIST arbitrary symbol handwritten dataset. Our model is flexible and can be easily extended to other languages. 

\bibliographystyle{plain}
\bibliography{egbib}
\end{document}